\pdfoutput=1

\documentclass[11pt]{article}

\usepackage[preprint]{acl}
\usepackage{graphicx}
\usepackage{subcaption}  
\usepackage{times}
\usepackage{latexsym}

\usepackage[utf8]{inputenc}

\usepackage{microtype}
\usepackage{colortbl}
\usepackage{xcolor}

\usepackage{inconsolata}

\usepackage{graphicx}
\usepackage{algorithm}
\usepackage{algorithmic}
\usepackage[most]{tcolorbox}
\usepackage{tabularx}
\usepackage{booktabs}
\usepackage{amsthm}
\usepackage{amsmath}
\usepackage{amssymb}
\usepackage{bm}
\usepackage{bbm}
\usepackage{multirow}
\title{\textit{Perception Compressor}: A Training-Free Prompt Compression Framework in Long Context Scenarios}

\author{
 \textbf{Jiwei Tang\textsuperscript{1}},
 \textbf{Jin Xu\textsuperscript{3}},
 \textbf{Tingwei Lu\textsuperscript{1}},
 \textbf{Zhicheng Zhang\textsuperscript{1}},
\\
 \textbf{Yiming Zhao\textsuperscript{4}},
 \textbf{Lin Hai\textsuperscript{1}},
 \textbf{Hai-Tao Zheng\textsuperscript{1,2}\thanks{Corresponding author: zheng.haitao@sz.tsinghua.edu.cn}},
\\
 \textsuperscript{1}Shenzhen International Graduate School, Tsinghua University \\
 \textsuperscript{2}Pengcheng Laboratory \\
 \textsuperscript{3}Ant Group \\
 \textsuperscript{4}School of Artificial Intelligence, Sun Yat-sen University \\
 \texttt{tangjw24@mails.tsinghua.edu.cn}
}

\begin{document}
\maketitle
\begin{abstract}
 Large language models (LLMs) demonstrate exceptional capabilities in various scenarios. However, they suffer from much redundant information and are sensitive to the position of key information in long context scenarios. To address these challenges, we present \textit{Perception Compressor}, a training-free prompt compression framework. It includes a perception retriever that leverages guiding questions and instruction to retrieve the most relevant demonstrations, a dual-slope ratio allocator to dynamically allocate compression ratios and open-book ratios, and a semi-guided iterative compression that retains key information at the token level while removing tokens that distract the LLM. We conduct extensive experiments on long context benchmarks, \textit{i.e.}, NaturalQuestions, LongBench, and MuSiQue. Experiment results show that \textit{Perception Compressor} outperforms existing methods by a large margin, achieving state-of-the-art performance. \footnote{Our code can be available at \url{https://github.com/Twilightaaa/PerceptionCompressor}.}
\end{abstract}

\section{Introduction}
Large language models (LLMs) (\textit{e.g.}, ChatGPT), known for their powerful generation and reasoning capabilities, have been widely applied in various scenarios. Commonly used methods like Chain-of-Thought (CoT)~\cite{wei2022chain}, In-Context Learning (ICL)~\cite{dong2022survey}, Retrieval Augment Generation (RAG)~\cite{Lewis_Perez_Piktus_Petroni_Karpukhin_Goyal_Küttler_Lewis_Yih_Rocktäschel_etal._2020}, effectively enhance the performance of LLMs by providing domain knowledge. However, these methods generally yield long prompts, even over thousands of tokens, \textit{i.e.}, long context. There are two challenges when using LLMs in long context scenarios: (1) Long prompts inherently contain much redundant information~\cite{shannon1951prediction, shi2023large}, and may exceed the window size of LLMs. \begin{figure}[thbp]
  \centering

  \begin{subfigure}[b]{0.9\linewidth}
    \centering
    \includegraphics[width=\linewidth]{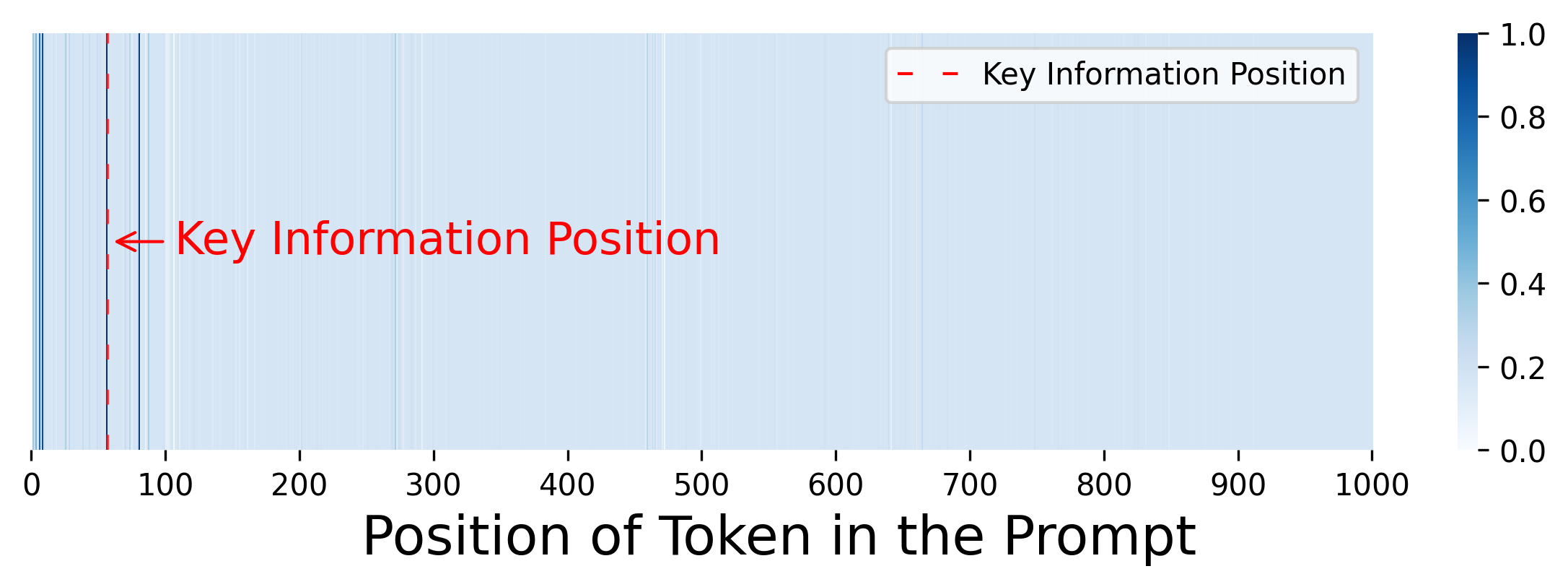}
    \caption{Normalized Contrast Perplexity}
    \label{fig:normalized_contrast_ppl}
  \end{subfigure}

  \begin{subfigure}[b]{0.9\linewidth}
    \centering
    \includegraphics[width=\linewidth]{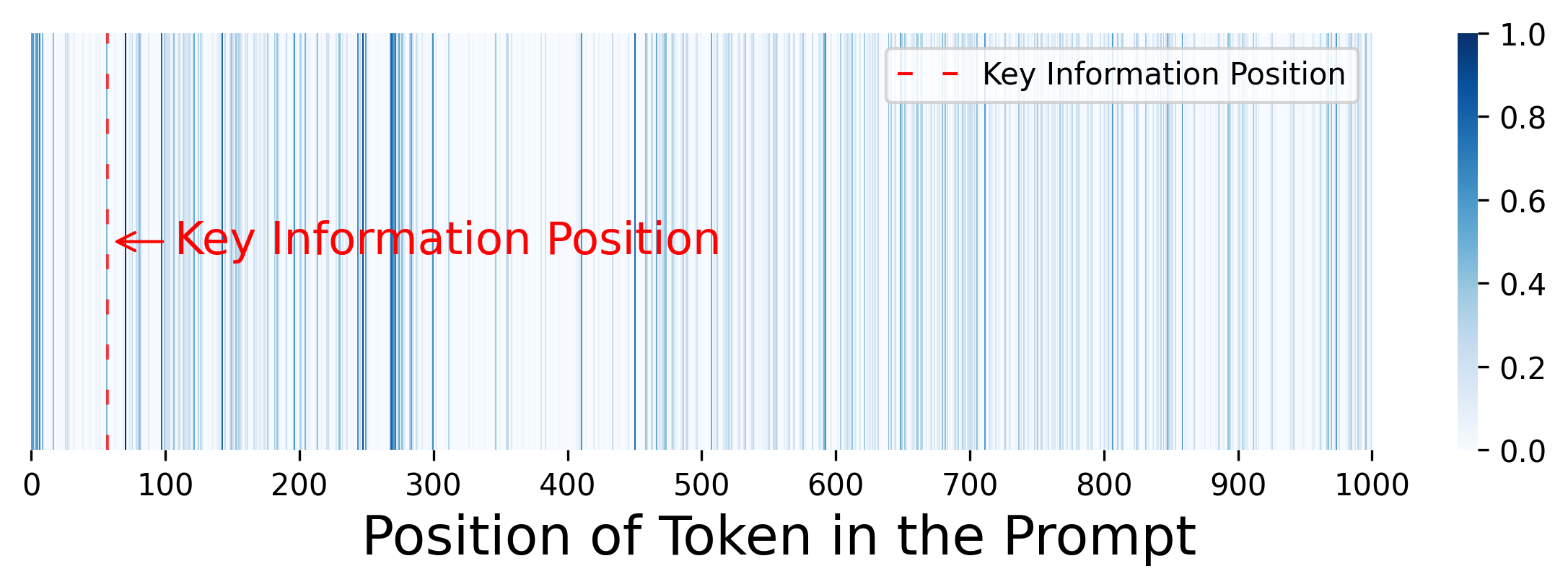}
    \caption{Normalized Perplexity}
    \label{fig:normalized_ppl}
  \end{subfigure}
  
  
  \caption{Perplexity v.s. Contrast Perplexity. Only a very small number of tokens related to key information have a high contrast perplexity, while the contrast perplexity of other tokens is nearly the same. However, the perplexity of different tokens varies significantly.}
  \label{fig:comparison_of_ppl}
\end{figure}(2) LLM is sensitive to the position of key information (relevant to the input question), \textit{i.e.}, the \textit{lost in the middle} challenge presented in~\citet{liu2024lost}. 

Some previous research~\cite{nijkamp2023xgen,han2023lm,peng2023yarn,sun2023retentive,ding2023longnet} extends the window size. However, while these methods increase the input token length, they struggle to overcome the notable decline in performance~\cite{ge2023context}. A prompt can be divided into different components (\textit{e.g.} instruction, demonstrations, and question). To address challenge one, other researchers focus on prompt compression methods. Among them, SelectiveContext~\cite{li2023compressing}, LLMLingua~\cite{jiang2023llmlingua}, and LLMLingua2~\cite{pan2024llmlingua} are task-agnostic methods, that ignore the input question during the compression process, potentially losing key information and failing to address challenge two in long context scenarios.

To enhance the density of key information and address challenge two, \citet{jiang2023longllmlingua} introduce LongLLMLingua. LLM can perceive the position of key information tokens (KITs) in prompt through its contrast perplexity~\cite{jiang2023longllmlingua}, \textit{i.e.}, tokens with high contrast perplexity contain key information (see Figure~\ref{fig:normalized_contrast_ppl}). LongLLMLingua first reorders and retrieves the demonstrations based on their relevance to the input question, then performs token-level compression based on contrast perplexity, removing tokens with lower contrast perplexity. 


However, this method has two limitations: (1) In the retrieval stage, this method not only ignores that the input question can be complex and difficult to understand, making it challenging to retrieve the most relevant demonstrations in high-noise long context scenarios but also neglects the impact of the instruction contained in the original prompt. The instruction contains all the guidance to generate answers that can significantly impact LLM's performance. (2) In the token-level compression stage, LongLLMLingua retains tokens with high contrast perplexity. While this preserves KITs with high contrast perplexity, it neglects the process of non-key information tokens (NITs). The number of NITs is significantly greater than that of KITs, and the perplexity of NITs is nearly the same (see Figure~\ref{fig:normalized_contrast_ppl}). This implies that once all KITs are retained, the selection of a large number of NITs is almost random. Irrelevant content can distract LLMs~\cite{shi2023large}. Randomly selecting NITs with vastly different perplexities (see Figure~\ref{fig:normalized_ppl}) can lead to the retention of distracting content, leading to much noise in the compressed prompt.




Inspired by these observations, we present \textit{Perception Compressor}, a training-free prompt compression framework, to address the challenges and limitations mentioned above. Specifically, we introduce a perception retriever that leverages instruction and guiding questions to retrieve demonstrations. Socratic Method is a philosophical inquiry technique that promotes deep understanding through questioning. This method is named after the ancient Greek philosopher Socrates, who was renowned for his dialogic teaching style. The essence of the Socratic Method lies in guiding students or participants to discover knowledge, clarify their thoughts, identify assumptions, and ultimately achieve a deeper level of understanding through a series of guiding questions~\cite{benson2011socratic}. We hope to activate the ``Scratic Thinking'' of LLMs through guiding questions, enabling them to identify underlying assumptions and prerequisites step-by-step, thus guiding them to focus on key information in high-noise long context scenarios. Next, we present a dual-slope ratio allocator to allocate all compression ratios and open-book ratios in the compression process. Knowledge with high perplexity is more uncertain for LLMs, hence tokens with high perplexity in NITs are more likely to distract the LLMs. Therefore, we utilize semi-guided iterative compression based on the open-book ratios to remove high perplexity tokens in NITs while retaining KITs. 

Our main contributions are four-fold: (1)We introduce a perception retriever, which achieves the highest recall@1 in the NaturalQuestions recall experiment (see Table~\ref{tab:recall}). And we use Bayes' Theorem to prove the rationality of perception retriever. (2)We present a dual-slope ratio allocator, which dynamically balances the compression ratios and the open-book ratios based on the varying relevance of each demonstration to the input question. (3)We propose semi-guided iterative compression to perform token-level compression, which removes NITs that distracts LLM while retaining KITs. (4)We conduct extensive experiments and comprehensive analysis on benchmarks in long context scenarios, \textit{i.e.}, NaturalQuestions, LongBench, and MuSiQue. The experiment results demonstrate the superiority and effectiveness of \textit{Perception Compressor}.

\begin{figure*}[thbp]
  \centering
  \includegraphics[width=1\textwidth]{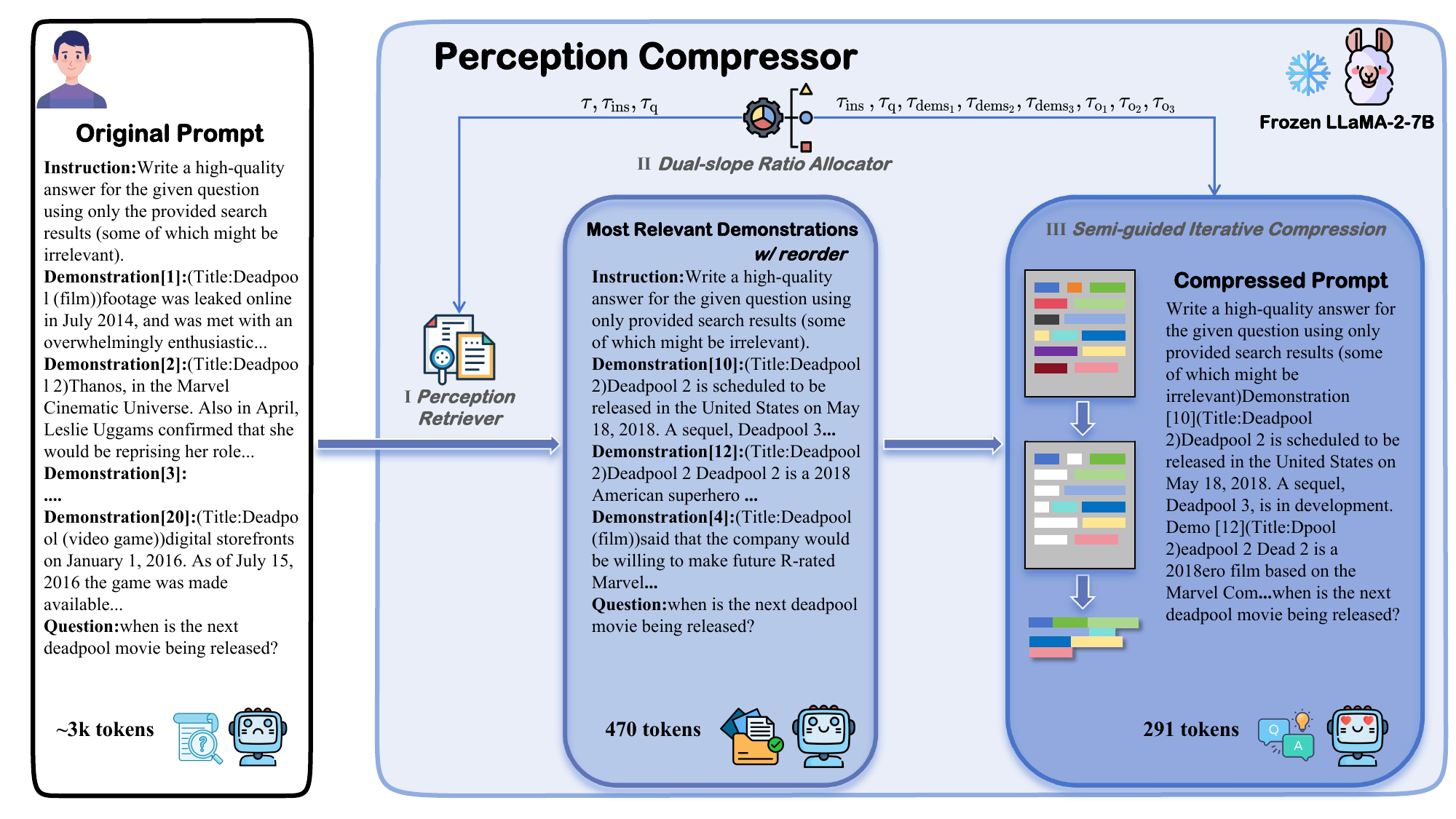}
  \caption{Framework of \textit{Perception Compressor}. The original prompt can be divided into instruction, demonstrations, and question. \textit{Perception Compressor} first uses the perception retriever to retrieve the most relevant demonstrations and reorders them from most to least relevant to the input question. Then, it performs a semi-guided iterative compression to obtain the final compressed prompt. The entire process is controlled by the compression ratios and open-book ratios allocated by the dual-slope ratio allocator.}
  \label{fig:framework}
  \vspace{-2mm}
\end{figure*}

\section{Related Work}
\subsection{Prompt Compression Methods}
As the input context gets longer, methods for compressing prompts receive widespread attention. Recently, prompt compression methods can mainly be divided into two categories: (1) Generating soft prompts. These methods~\cite{li2024500xcompressor,Cao2024RetainingKI,rau2024context,chuang2024learning,Chevalier2023AdaptingLM,zhang2024compressing} mainly focus on learning soft tokens to reduce the length of input tokens, which requires further parameter fine-tuning of LLM, long training time, and high computational cost. Moreover, LLMs trained by these methods are generally domain-specific, with relatively poor generalization. (2) Explicitly compress prompt. 
Recently, some methods based on information theory have emerged, such as Selective Context~\cite{li2023compressing}, LLMLingua~\cite{jiang2023llmlingua}. To use bidirectional context, \citet{pan2024llmlingua} introduce LLMLingua2. However, these methods ignore the relevance of the content to the input question. To increase the density of key information, \citet{jiang2023longllmlingua} propose LongLLMLingua, which firstly reorders the demonstrations based on their relevance to the input question, keeping the most relevant demonstrations. Then use contrast perplexity ITPC to perform token-level compression. Another common method is to generate a summary of the given context, which requires further training of the LLM~\cite{xu2023recomp,wang2023learning}.

\subsection{Extending Window Size}
Recently, a series of studies have been proposed for extending context window size. This extension has been pursued through various avenues. One strategy is staged pre-training~\cite{nijkamp2023xgen}, where the context window size gradually increases over successive pre-training stages. 
Some studies involve interpolating position embeddings~\cite{han2023lm,peng2023yarn}, aiming to adapt these embeddings to accommodate longer context.
Furthermore, researchers have delved into adjusting attention mechanisms to handle expansive context windows more efficiently. This includes exploring linear or sparse attention mechanisms~\cite{ding2023longnet,sun2023retentive}. Additionally, there is a line of study that revolves around leveraging external memory modules for storing context~\cite{Bertsch_Alon_Neubig_Gormley_2023,tworkowski2024focused}. These modules alleviate the burden on the LLM by offloading context information to dedicated memory units.

However, while these methods offer promising avenues for extending the context window of LLMs, they generally lead to relatively inferior performance. 

\subsection{Retrieval Methods}
The retrieval methods can be divided into two categories: (1) Sparse retrieval. Sparse retrieval retrieves the most similar items from an index based on query keywords or feature vectors, such as BM25. (2) Dense retrieval. Dense retrieval methods~\cite{bge_embedding,reimers2019sentence,jiang-etal-2023-low,günther2023jina} involve training retrieval models on large corpora, generating dense vector representations for queries and demonstrations through model inference, and computing similarities based on these vector representations.
\begin{table*}[htb]
\centering
\footnotesize
\setlength{\tabcolsep}{8 pt}
\resizebox{\textwidth}{!}{%
\begin{tabular}{lcccccccccccccccccccc}
\toprule
Methods & \multicolumn{20}{c}{Recall} \\ 
\cmidrule{2-21} 
        & @1    & @2    & @3    & @4    & @5    & @6    & @7    & @8    & @9    & @10   & @11   & @12   & @13   & @14   & @15   & @16   & @17   & @18   & @19   & @20   \\ 
\midrule
LLMLingua   & 4.4 & 7.1 & 9.3 & 12.2 & 14.8 & 17.6 & 20.6 & 23.0 & 25.9 & 29.3 & 32.6 & 36.1 & 39.5 & 43.2 & 47.4 & 52.7 & 58.4 & 66.3 & 77.6 & 100.0 \\
BM25        & 8.0 & 13.9 & 19.6 & 24.6 & 29.4 & 34.2 & 38.5 & 42.7 & 46.7 & 52.1 & 56.4 & 60.9 & 64.9 & 69.2 & 73.5 & 77.7 & 82.6 & 87.3 & 92.7 & 100.0 \\ 
Bge         & 35.6 & 49.5 & 57.8 & 63.8 & 69.0 & 72.9 & 76.4 & 79.4 & 82.1 & 84.4 & 86.3 & 88.2 & 89.6 & 91.1 & 92.4 & 93.8 & 95.3 & 96.7 & 98.2 & 100.0 \\
Gzip        & 50.1 & 55.7 & 59.7 & 63.0 & 65.8 & 68.5 & 70.6 & 72.8 & 74.7 & 77.2 & 79.8 & 81.7 & 84.1 & 85.9 & 87.6 & 89.3 & 91.3 & 93.3 & 95.6 & 100.0 \\
Jina       & 52.2 & 66.0 & 74.0 & 78.6 & 82.2 & 85.4 & 87.6 & 89.2 & 91.0 & 92.2 & 93.3 & 94.5 & 95.4 & 96.5 & 97.0 & 97.6 & 98.4 & 99.0 & 99.4 & 100.0 \\
OpenAI Embedding     & 52.9 & 67.0 & 75.9 & 81.5 & 85.7 & 88.3 & 90.6 & 92.1 & 93.2 & 94.1 & 95.4 & 96.3 & 97.1 & 97.8 & 98.2 & 98.5 & 98.8 & 99.2 & 99.6 & 100.0 \\
SentenceBert & 54.7 & 66.6 & 73.0 & 77.5 & 81.3 & 84.3 & 86.5 & 88.1 & 89.7 & 91.5 & 92.8 & 93.9 & 94.9 & 95.5 & 96.2 & 96.8 & 97.5 & 98.3 & 98.9 & 100.0 \\
BgeLLMembedder & 59.9 & 73.3 & 80.3 & 85.1 & 87.8 & 89.9 & 91.5 & 92.8 & 93.6 & 94.6 & 95.7 & 96.5 & 97.2 & 97.6 & 98.1 & 98.4 & 98.7 & 99.0 & 99.6 & 100.0 \\
BgeReranker & 62.9 & 74.5 & 80.0 & 83.4 & 85.8 & 88.2 & 89.8 & 91.5 & 92.8 & 93.9 & 94.9 & 95.9 & 96.5 & 97.5 & 97.9 & 98.2 & 98.8 & 99.3 & 99.5 & 100.0 \\
LongLLMLingua $r_k$ & 67.1 & 78.1 & 82.9 & 86.0 & 88.7 & 90.5 & 91.6 & 93.0 & 94.2 & 95.4 & 96.3 & 97.1 & 97.6 & \textbf{98.2} & \textbf{98.8} & 99.0 & \textbf{99.4} & \textbf{99.7} & \textbf{99.9} & 100.0 \\ 
\midrule
{\cellcolor[rgb]{0.925,0.957,1}}\textbf{Perception retriever} & {\cellcolor[rgb]{0.925,0.957,1}}\textbf{72.3} & {\cellcolor[rgb]{0.925,0.957,1}}\textbf{81.7} & {\cellcolor[rgb]{0.925,0.957,1}}\textbf{86.1} & {\cellcolor[rgb]{0.925,0.957,1}}\textbf{88.3} & {\cellcolor[rgb]{0.925,0.957,1}}\textbf{90.2} & {\cellcolor[rgb]{0.925,0.957,1}}\textbf{91.7} & {\cellcolor[rgb]{0.925,0.957,1}}\textbf{92.9} & {\cellcolor[rgb]{0.925,0.957,1}}\textbf{94.2} & {\cellcolor[rgb]{0.925,0.957,1}}\textbf{95.0} & {\cellcolor[rgb]{0.925,0.957,1}}\textbf{95.8} & {\cellcolor[rgb]{0.925,0.957,1}}\textbf{96.4} & {\cellcolor[rgb]{0.925,0.957,1}}\textbf{97.2} & {\cellcolor[rgb]{0.925,0.957,1}}\textbf{97.6} & {\cellcolor[rgb]{0.925,0.957,1}}98.1 & {\cellcolor[rgb]{0.925,0.957,1}}98.5 & {\cellcolor[rgb]{0.925,0.957,1}}\textbf{99.0} & {\cellcolor[rgb]{0.925,0.957,1}}99.2 & {\cellcolor[rgb]{0.925,0.957,1}}99.6 & {\cellcolor[rgb]{0.925,0.957,1}}99.8 & {\cellcolor[rgb]{0.925,0.957,1}}\textbf{100.0} \\ 
\bottomrule

\end{tabular}
}
\caption{Comparison of recall on NaturalQuestions (20 documents)~\cite{liu2024lost}. We use the recall rate of ground truth document as the evaluation metric.}
\label{tab:recall}
\vspace{-3mm}
\end{table*}

\section{Task Formulation}
Given a original prompt with augmented context $\mathbf{x}=(\mathbf{x}^{\text{ins}}, \mathbf{x}^{\text{dems}_{1}}, ..., \mathbf{x}^{\text{dems}_{N}}, \mathbf{x}^{\text{q}})$, which consists of the instruction $\mathbf{x}^{\text{ins}}$, $N$ demonstrations $\left\{\mathbf{x}^{\text{dems}_{i}}\right\}_{i=1}^{N}$, and the input question $\mathbf{x}^{\text{q}}$. A compression ratio $1/\tau$, $\tau \in [0,1]$. The objective of prompt compression can be formulated as:
\begin{equation}
    \min _{\widetilde{x},\tau} d\left(p\left(\widetilde{\mathbf{y}} \mid \widetilde{\mathbf{x}}\right), p\left(\mathbf{y} \mid \mathbf{x}\right)\right)
\end{equation}

where $d(\cdot,\cdot)$ is a function measuring the distance between two distributions (\textit{e.g.}, KL divergence); $\widetilde{y}$ represents the LLM-generated results derived by compressed prompt $\widetilde{x}$ and $y$ denotes the ground-truth answer. The distribution of $\widetilde{y}$ is expected to be as similar to $y$ as possible.

\section{Method}
In this section, we elaborate on our proposed prompt compression framework, \textit{i.e.}, \textit{Perception Compressor}. The overall framework and entire process are shown in Figure~\ref{fig:framework}.

\subsection{Perception Retriever}
\label{sec:perception retriever}
\paragraph{Generate Guiding Questions}
We first employ an LLM (\textit{e.g.}, ChatGPT) to generate guiding questions based on the input question. Specifically, given a question, we prompt the LLM to generate a set of guiding questions. 
\begin{equation}
    LLM(q_0) = \{q_1, q_2, \ldots, q_n\}
\end{equation}

where we use prompt\footnote{Specifically, ``\textit{Please provide \textbf{n} most helpful guiding questions to address the original question: \{original question\}"}} to guide LLM in generating guiding questions; $q_0$ is the input question; $\{q_1, q_2, \ldots, q_n\}$ is the generated $n$ guiding questions. 
\paragraph{Calculate Semantic Similarity}
We use SentenceBert to calculate the semantic similarity between $q_0$ and $q_0, ..., q_n$.
\begin{equation}
    e_i = F(q_i)
\end{equation}

where $F(\cdot)$ is semantic feature extractor, \textit{i.e.}, SentenceBert; $e_i$, where $i \in \{0, 1, \ldots, n\}$, is the semantic feature vector corresponding to question $q_i$. Then we can define the semantic similarity $w_i$ between $q_{0}$ and $q_{i}$ as:
\begin{equation}
    w_i = \frac{e_0 \cdot e_i}{\|e_0\| \cdot \|e_i\|}
\end{equation}

\paragraph{Perception Perplexity}
We first calculate the condition perplexity $r_{k,j}$ for each demonstration $\mathbf{x}^{\text{dems}_{k}}, k \in \{1, \ldots, N\}$, where $N$ is the number of all demonstrations.

\begin{equation}
\label{eq:perception_perlexity}
\begin{aligned}
    r_{k,j} = &\sum_{i=1}^{L_{ins} + L_{q_j} + L_{r}} g\left( \mathbf{x}^{\text{con}_j}_i \right) \log p\left( \mathbf{x}^{\text{con}_j}_i \mid \mathbf{x}^{\text{dems}_{k}} \right)
\end{aligned}
\end{equation}

where $g(\cdot)$ represents the probability distribution of the ground truth; $\mathbf{x}^{\text{con}_j}$ is the concatenation of $\mathbf{x}^{\text{ins}}$, $\mathbf{x}^{\text{q}_{j}}$ and $\mathbf{x}^{\text{r}}$; $\mathbf{x}^{\text{con}_j}_i$ denotes the $i$-th token in $\mathbf{x}^{\text{con}_j}$; $\mathbf{x}^{\text{dems}_{k}}$ and $\mathbf{x}^{\text{r}}$ refer the $k$-th demonstration and regularization constraint\footnote{For fair comparison, We use the same regularization constraint in \citet{jiang2023longllmlingua} to strengthen the connection between $\mathbf{x}^{\text{dems}_{k}}$ and $\mathbf{x}^{\text{q}}$.}, respectively. \par
Then, we can obtain perception perplexity $r_k$ of demonstration $\mathbf{x}^{\text{dems}_{k}}$.
\begin{equation}
\label{eq:ws_perception_perlexity}
    r_k = \sum_{j=0}^{n} w_j \cdot r_{k,j}
\end{equation}

Through the derivation in Appendix~\ref{apx:derivation}, we can obtain:
\begin{equation}
\label{eq:perception_bayes}
\begin{aligned}
\{r_k\}_i \propto \log \prod_{j=0}^{n}  {p\left(\mathbf{x}^{\text{dems}_{k}} \mid \mathbf{x}^{\text{con}_j}_i \right)}^{w_j}
\end{aligned} 
\end{equation}


where $\{r_{k}\}_i$ is the impact of the $i$-th token in $\mathbf{x}^{\text{con}_j}$ on the perception perplexity. 

Observing Equation~\eqref{eq:perception_bayes}, we can derive two insights: (1)$w_j$ is a scaling factor. The more relevant $q_j$ is to the $q_0$, and the greater its impact on $\{r_{k}\}_i$. (2)$\{r_k\}_i$ is proportional to the log-likelihood function~\cite{fisher1922mathematical} of the conditional probability given by $\mathbf{x}^{\text{con}_j}_i$, where $j$ ranges from $0$ to $n$. This indicates that $\{r_k\}_i$ reflects the relevance of the current demonstration $\mathbf{x}^{\text{dems}_k}$ to the entire set of $\mathbf{x}^{\text{con}_j}_i$. The larger $\{r_k\}_i$ is, the stronger the relevance.

\paragraph{Reorder the Demonstrations} We reorder the demonstrations based on perception perplexity, from high to low. We retain $\mathbf{x}^{\text{dems}_{k}}$ with higher $r_k$ as retrieval results until the total token length meets the compression ratio constraint.

\subsection{Dual-slope Ratio Allocator}
\label{sec:allocator}
Different components of the prompt (\textit{e.g.} instruction, demonstrations, question) have varying degrees of redundancy. Obviously, the demonstrations have the highest degree of redundancy, so higher compression ratios should be allocated to them. 

We need to predefine $\tau$, $\tau_{\text{ins}}$ and $\tau_{\text{q}}$ and set the coarse-grained control coefficient $\mu$ to determine the number of retrieved demonstrations. Then, we can get the basic compression ratio for each demonstration.

\begin{equation}
\label{eq:cal tau dems}
    \tau_{\text{dems}} =\frac{L^{'}_{\text{dems}}-(\mu \tau L-\tau_{\text {ins }} L_{\text {ins }}-\tau_{\text {q }} L_{\text {q }})}{ L^{'}_{\text {dems }}}
\end{equation}

where $L$ is the token length of the original prompt; $L^{'}_{\text{docs}}$ denotes the token length of all retained demonstrations; $L_{\text{q}}$ and $L_{\text{ins}}$ are the token length of the input question and instruction, respectively.

To assign a lower compression ratio to demonstrations that are more relevant to the input question, we introduce the first slope $k_1$.
\begin{equation}
\small
\tau_{\text{dems}_{k}} = max(min(1, \tau_{\text{dems}} + (1 - \frac{2\text{rank}(r_k)}{N_d})\cdot k_{1}) , 0)
\end{equation}

where $k_1 > 0$; $rank(\cdot)$ is the ranking index of perception perplexity $r_k$ (\textit{e.g.}, 0, 1);$N_d$ represents the number of demonstrations retained by the Perception retriever.


$\tau_o$ is the open-book ratio, which is the budget for considering the contrast perplexity. When the compression ratio of a demonstration is high, we should prioritize the preservation of key information with high contrast perplexity. Therefore, we introduce $k_2$.

\begin{equation}
\small
\tau_{\text{o}_{k}} = max(min(1, \tau_{\text{o}} - (1 - \frac{2\text{rank}(r_k)}{N_d})\cdot k_{2}) , 0)
\end{equation}

where $k_2 > 0$.

\subsection{Semi-guided Iterative Compression}

Because LLM's context window size is of limited length, and perplexity may model local information instead of catching long-range dependency~\cite{hu2024perplexityreflectlargelanguage}, we divide the complete context into several segments $s=\{s_1,s_2,...s_m\}$ and then compress them sequentially from front to back.

We use conditional probability modeling for compression, which can be formulated as:

\begin{equation}
\begin{aligned}
p(\bm{\widetilde{s}}_j) \approx \prod_{i=1}^{L_{s,j} + \sum_k^{j-1}\widetilde{L}_{s,k}} p(s_{j,i}|\widetilde{s}_{<j},s_{j,<i})
\end{aligned}
\label{eq:prompt_ppl_iterative}
\end{equation}

where $s_{j,i}$ denotes the $i$-th token in the $j$-th segment; $L_{s, j}$ and $\widetilde{L}_{s, j}$ represent the token length of $j$-th original and compressed segment, respectively.


The conditional contrast perplexity for each segment is:
\begin{equation}
\label{eq:contrast perplexity}
\small
\begin{aligned}
Q(s_{j,i}) &= g(s_{j,i}) \log p(s_{j,i} \mid q_{0}, \widetilde{s}_{<j}, s_{j,<i})  \\
&\quad  - g(s_{j,i}) \log p(s_{j,i} \mid \widetilde{s}_{<j}, s_{j,<i}) \\
&= g(s_{j,i}) \log \frac{p(s_{j,i} \mid q_{0}, \widetilde{s}_{<j}, s_{j,<i})}{p(s_{j,i} \mid \widetilde{s}_{<j}, s_{j,<i})}
\end{aligned} 
\end{equation}





The conditional perplexity can be formulated as:
\begin{equation}
    P(s_{j,i}) = g(s_{j,i}) \log p(s_{j,i} \mid \widetilde{s}_{<j},s_{j,<i})
\end{equation}

When the contrast perplexity $Q(s_j)$ and perplexity $P(s_j)$ for each token in segment are obtained, the thresholds $\gamma^{Q}_j$ and $\gamma^{P}_j$ are dynamically calculated based on the retention ratios $\tau_{\bm{s}_j}*\tau_{o}$ and $\tau_{\bm{s}_j}*(1-\tau_{o})$, respectively.



\begin{equation}
\tau_{\bm{s}_j} =\left\{
\begin{aligned}
\tau_{\text{ins}}, &\quad \text{if $\bm{s}_j \in \bm{x}^{\text{ins}}$}, \\
\tau_{\text{dems}}, &\quad \text{if $\bm{s}_j \in \bm{x}^{\text{dems}}$},\\
\tau_{\text{q}} , &\quad \text{if $\bm{s}_j \in \bm{x}^{\text{q}}$}.\\
\end{aligned}
\right.
\label{eq:tau_for_gamma}
\end{equation}



First, we retain KITs through Equation~\eqref{eq:retain}. 
\begin{equation}
\label{eq:retain}
\widetilde{s}^{K}_{j} = \{ s_{j, i} \mid \, Q(s_{j, i}) \geq \gamma^{Q}_{j} \}   
\end{equation}

where $\widetilde{s}^{K}_{j}$ is the retained KITs in $\widetilde{s}_{j}$.

Then, we remove irrelevant supplementary knowledge that affects the performance of LLMs in NITs via Equation~\eqref{eq:delete}.
\begin{equation}
\label{eq:delete}
\widetilde{s}^{N}_{j} = \{ Q(s_{j, i}) < \gamma^{Q}_{j} \text{ and } P(s_{j, i}) \leq \gamma^{P}_{j} \}    
\end{equation}

where $\widetilde{s}^{N}_{j}$ is the retained NITs in $\widetilde{s}_{j}$.

Therefore, we can get $\widetilde{s}_{j}$:
\begin{equation}
\widetilde{s}_j = \widetilde{s}^{K}_{j} \cup \widetilde{s}^{N}_{j}   
\end{equation}


The compressed prompt $\widetilde{s}$ is formed by sequentially concatenating each compressed segment.
\begin{equation}
    \widetilde{s} = \widetilde{s}_{1} \odot \widetilde{s}_{2} \odot \ldots \odot \widetilde{s}_{m}
\end{equation}

\begin{table*}[tb]
    \small
    \centering
    \setlength{\tabcolsep}{4.6 mm}
    \resizebox{\textwidth}{!}{%
    \begin{tabular}{l|ccccc|ccccc|cc}
    \toprule
        \multirow{2}{*}{Methods} &  \multicolumn{5}{@{}c}{{\bf LongChat-7B-v1.5-32k}} &  \multicolumn{5}{@{}c}{{\bf LLaMA-3-8B-Instruct}} & \multicolumn{2}{@{}c}{{\bf Length}} \\
        & 1st & 5th & 10th & 15th & 20th  & 1st & 5th & 10th & 15th & 20th  & Tokens & $1/\tau$\\
    \midrule
    \midrule
    \multicolumn{13}{@{}c}{{ \textit{2x constraint}}} \\
    \midrule
      \multicolumn{13}{@{}l}{{ \textit{Retrieval-based Methods}}} \\ 
OpenAI             &      62.0     &    61.7       &     61.4      &      61.2    &  60.8     &   73.0   &  73.5  &  73.0 &  74.1  &  73.6 &        1,408              &                 2.1x          \\
SentenceBert             & 61.1          & 60.3          & 60.5          & 59.9          & 59.8     &   72.8   &  73.4  &  72.8  &  73.0  & 72.9     & 1,410 &     2.1x                      \\
BgeReranker            & 61.9          & 61.1          & 60.7          & 61.3          & 59.6   &   74.5   &  74.1  &  74.7 &  73.8  &  72.6      & 1,405 &     2.1x                      \\
BgeLLMembedder           & 62.5          & 61.8          & 62.2          & 61.6          & 60.8    &   74.0   &  73.7  &  74.2 &  74.0  &  74.0     & 1,407 &     2.1x                      \\
    \cmidrule (r){1-1}\cmidrule (lr){2-6} \cmidrule (lr){7-11} \cmidrule (lr){12-13}
    \multicolumn{13}{@{}l}{{ \textit{Compression-based Methods}}} \\
SelectiveContext          & 47.2          & 41.8          & 40.5          & 39.2          & 41.3    &   64.1   &  58.3  &  57.2 &  56.6  &  55.9      & 1,725 &    1.7x                       \\
LLMLingua                &     39.7      &     37.4      &    35.8       &    34.7       & 36.0   &  48.1    &  46.7  &  46.0 &  44.3  &  44.4      &    1,535       &     1.9x                      \\
LLMLingua2             &    54.8       &      46.4     &      44.0     &     42.5     &  44.3     &  70.8    &  62.7  & 62.1  &  61.5  & 61.8    &      1,475              &                 2.0x          \\
LongLLMLingua            & 62.3          & 61.7         & 62.4         & 62.4          & 61.8   &   77.0   &  76.0  &  74.8 &  75.2  &  74.9     & 1,444 &             2.1x              \\ 
    \cmidrule (r){1-1}\cmidrule (lr){2-6} \cmidrule (lr){7-11} \cmidrule (lr){12-13}
    {\cellcolor[rgb]{0.925,0.957,1}}\textbf{Perception Compressor} & {\cellcolor[rgb]{0.925,0.957,1}}\textbf{64.7} & {\cellcolor[rgb]{0.925,0.957,1}}\textbf{64.1} & {\cellcolor[rgb]{0.925,0.957,1}}\textbf{63.5} & {\cellcolor[rgb]{0.925,0.957,1}}\textbf{63.2} & {\cellcolor[rgb]{0.925,0.957,1}}\textbf{64.6} & {\cellcolor[rgb]{0.925,0.957,1}}\textbf{79.5} &  {\cellcolor[rgb]{0.925,0.957,1}}\textbf{78.3} & {\cellcolor[rgb]{0.925,0.957,1}}\textbf{78.6} & {\cellcolor[rgb]{0.925,0.957,1}}\textbf{79.1} & {\cellcolor[rgb]{0.925,0.957,1}}\textbf{79.1} & {\cellcolor[rgb]{0.925,0.957,1}}1,373 & {\cellcolor[rgb]{0.925,0.957,1}}2.1x \\

    \midrule
    \midrule
    \multicolumn{13}{@{}c}{{ \textit{4x constraint}}} \\
    \midrule
      \multicolumn{13}{@{}l}{{ \textit{Retrieval-based Methods}}} \\ 
OpenAI            &     60.5      &      61.4     &      60.5     &    60.8      &  60.7    &   71.1   &  70.8  & 72.7  &  72.4  & 72.1      &          664          &         4.4x                  \\
SentenceBert            & 60.0          & 59.6          & 60.1          & 59.5          & 58.9  &   72.0   &  72.1  &  71.7  &  71.8   &  72.5        &    665                  &     4.4x                      \\
BgeReranker            & 62.9          & 62.4          & 62.7          & 61.6          & 59.9  &   74.7   &  75.0   & 74.8  &  74.8  &  72.3        &      664                &      4.4x                     \\
BgeLLMembedder          & 62.5          & 57.7          & 61.6          & 61.2          & 61.2   &   73.5   &  72.9  &  73.5 &  73.1  &  74.3       &       664               &        4.4x                   \\
    \cmidrule (r){1-1}\cmidrule (lr){2-6} \cmidrule (lr){7-11} \cmidrule (lr){12-13}
    \multicolumn{13}{@{}l}{{ \textit{Compression-based Methods}}} \\
    SelectiveContext         & 34.0          & 31.1          & 30.3          & 29.0          & 31.0   &   47.6   &  46.7  & 44.8  &  44.5  &  43.2       &          828              &      3.6x                       \\
LLMLingua                &     30.4      &      29.5     &    29.2      &      28.5     & 28.9  &   38.2   &  38.0  & 37.1  &  35.8  & 35.6        &           764           &         3.9x                  \\
LLMLingua2           &       42.9    &     35.7      &    33.9       &  31.3       &  33.5   &   59.3   &  53.6  &  52.5  &  50.1  & 48.9      &       741              &        4.0x                   \\
LongLLMLingua          & 63.4          & 63.4         & 62.9          & 62.4          & 63.0    &   76.5   &  75.0  & 74.8  &  74.8 &  75.0      &         736             &       4.0x                    \\
    \cmidrule (r){1-1}\cmidrule (lr){2-6} \cmidrule (lr){7-11} \cmidrule (lr){12-13}
    {\cellcolor[rgb]{0.925,0.957,1}}\textbf{Perception Compressor} & {\cellcolor[rgb]{0.925,0.957,1}}\textbf{66.3} & {\cellcolor[rgb]{0.925,0.957,1}}\textbf{66.0} & {\cellcolor[rgb]{0.925,0.957,1}}\textbf{66.0} & {\cellcolor[rgb]{0.925,0.957,1}}\textbf{66.1} & {\cellcolor[rgb]{0.925,0.957,1}}\textbf{66.2} &  {\cellcolor[rgb]{0.925,0.957,1}}\textbf{79.6} & {\cellcolor[rgb]{0.925,0.957,1}}\textbf{79.5} & {\cellcolor[rgb]{0.925,0.957,1}}\textbf{80.7} & {\cellcolor[rgb]{0.925,0.957,1}}\textbf{80.2} & {\cellcolor[rgb]{0.925,0.957,1}}\textbf{80.0} & {\cellcolor[rgb]{0.925,0.957,1}}697 & {\cellcolor[rgb]{0.925,0.957,1}}4.2x \\
    \midrule
    \midrule
    Original Prompt & 63.1 & 56.6 & 53.1 & 53.1 & 56.1 &  76.6 & 67.5 & 65.8 & 67.4 & 65.6 & 2,949 & -  \\
    \cmidrule (r){1-1}\cmidrule (lr){2-6} \cmidrule (lr){7-11} \cmidrule (lr){12-13}
    Zero-shot &  & & 31.0 & & & & & 46.7 && & 10 & 294x \\
    \bottomrule
    \end{tabular}%
    }
    \caption{Performance of different methods under different target compression constraints on NaturalQuestions. 1st, 5th, 10th, 15th, and 20th refer to the positions of the document containing ground truth among the all 20 documents.}
    \vspace{-2mm}
    \label{tab:nq}
\end{table*}

\begin{table*}[tb]
    \small
    \centering
    \setlength{\tabcolsep}{4.6 mm}
    \resizebox{1\textwidth}{!}{%
    \begin{tabular}{l|ccccccc|cc}
    \toprule
        \multirow{2}{*}{Methods} &  \multicolumn{7}{@{}c}{{\bf LongBench}}  & \multicolumn{2}{@{}c}{{\bf Length}} \\
        & SingleDoc & MultiDoc & Summ. & Code& Synth. & FewShot & Avg & Tokens & $1/\tau$ \\
    \midrule
    \midrule
    \multicolumn{10}{@{}c}{{ \textit{3,000 tokens constraint}}} \\
    \midrule
      \multicolumn{10}{@{}l}{{ \textit{Retrieval-based Methods}}} \\ 
    SentenceBert & 24.0 & 23.0 & 26.0 & 37.7 & 32.9 & 64.1 & 34.6 & 3,001 & 3.4x  \\
    BgeReranker & 24.4 & 25.1 & 25.9 & 38.4 & 34.7 & 64.8 & 35.6 & 3,001 & 3.4x \\
    BgeLLMembedder & 24.3 & 25.3 & 25.7 & 37.3 & 27.5 & 65.1 & 34.2 & 3,001 & 3.4x  \\
    \cmidrule (r){1-1}\cmidrule (lr){2-10} 
    \multicolumn{7}{@{}l}{{ \textit{Compression-based Methods}}} \\
    SelectiveContext & 21.1 & 22.7 & 24.8 & 44.8 & 11.9 & 49.2 & 29.1 & 3,065 & 3.4x  \\
    LLMLingua & 22.3 & 16.2 & 25.0 & 49.0 & 12.6 & 65.7 & 32.1 & 3,004 & 3.4x  \\
    LLMLingua2 & 24.4 & 24.8 & 25.7 & 41.8 & 25.2 & 48.7 & 31.8 & 3,166 & 3.2x  \\
    LongLLMLingua & 23.6 & 27.8 & 25.6 & 43.1 & 52.3 & 65.9 & 39.7 & 3,051 & 3.4x  \\
    \cmidrule (r){1-1}\cmidrule (lr){2-10} 
    {\cellcolor[rgb]{0.925,0.957,1}}\textbf{Perception Compressor} & {\cellcolor[rgb]{0.925,0.957,1}}\textbf{27.6} & {\cellcolor[rgb]{0.925,0.957,1}}\textbf{28.5} & {\cellcolor[rgb]{0.925,0.957,1}}\textbf{26.1} & {\cellcolor[rgb]{0.925,0.957,1}}\textbf{50.0} & {\cellcolor[rgb]{0.925,0.957,1}}\textbf{54.2} & {\cellcolor[rgb]{0.925,0.957,1}}\textbf{66.5} & {\cellcolor[rgb]{0.925,0.957,1}}\textbf{42.2} & {\cellcolor[rgb]{0.925,0.957,1}}2,840 & {\cellcolor[rgb]{0.925,0.957,1}}3.6x \\

    \midrule
    \midrule
    \multicolumn{10}{@{}c}{{ \textit{2,000 tokens constraint}}} \\
    \midrule
      \multicolumn{10}{@{}l}{{ \textit{Retrieval-based Methods}}} \\ 
    SentenceBert & 25.6 & 25.5 & 26.0 & 39.5 & 30.0 & 64.0 & 35.1 & 2,244 & 4.6x  \\
    BgeReranker & 24.7 & 25.4 & 25.4 & 39.6 & 32.9 & 63.2 & 35.2 & 2,247 & 4.6x \\
    BgeLLMembedder & 23.1 & 26.2 & 25.7 & 39.4 & 22.2 & 63.3 & 33.3 & 2,245 & 4.6x  \\
    \cmidrule (r){1-1}\cmidrule (lr){2-10} 
    \multicolumn{7}{@{}l}{{ \textit{Compression-based Methods}}} \\
    SelectiveContext & 19.8 & 22.2 & 23.1 & 42.8 & 8.2 & 47.2 & 27.2 & 2,273 & 4.5x  \\
    LLMLingua & 20.4 & 14.4 & 24.2 & 47.0 & 9.2 & 65.1 & 30.8 & 2,251 & 4.6x  \\
    LLMLingua2 & 25.2 & 25.8 & 25.3 & 40.5 & 18.3 & 46.1 & 30.2 & 2,329 & 4.4x  \\
    LongLLMLingua & 24.1 & 25.8 & 25.2 & 44.3 & 38.7 & 65.3 & 37.2 & 2,103 & 4.9x  \\
    \cmidrule (r){1-1}\cmidrule (lr){2-10} 
    {\cellcolor[rgb]{0.925,0.957,1}}\textbf{Perception Compressor} & {\cellcolor[rgb]{0.925,0.957,1}}\textbf{26.6} & {\cellcolor[rgb]{0.925,0.957,1}}\textbf{26.5} & {\cellcolor[rgb]{0.925,0.957,1}}\textbf{26.2} & {\cellcolor[rgb]{0.925,0.957,1}}\textbf{54.2} & {\cellcolor[rgb]{0.925,0.957,1}}\textbf{51.7} & {\cellcolor[rgb]{0.925,0.957,1}}\textbf{65.9} & {\cellcolor[rgb]{0.925,0.957,1}}\textbf{41.9} & {\cellcolor[rgb]{0.925,0.957,1}}1,896 & {\cellcolor[rgb]{0.925,0.957,1}}5.4x \\
    \midrule
    \midrule
    Original Prompt & 28.6 & 21.5 & 16.0 & 55.6 & 37.0 & 65.6 & 37.4 & 10,276 & -  \\

    \bottomrule
    \end{tabular}
    }
    \caption{Performance of different methods under different tokens constraints on LongBench~\cite{bai2023longbench}.}
    \vspace{-2mm}
    \label{tab:longbench}
\end{table*}

\section{Experiments}
In this section, We seek to answer the following research questions: (1)How does the Perception Compressor address the two challenges in long context scenarios, \textit{i.e.}, sensitivity to the position of key information, and excessively long input sequences? (2)How does Perception Compressor compare to baseline methods in long context benchmarks? (3)What is the impact of each component within the Perception Compressor on the overall method?

We also conduct some supplementary experiments, including parameter sensitivity analysis (see Appendix~\ref{apx:psa}), latency and memory usage (see Appendix~\ref{apx:latency_and_mem}), evaluation on black-box large models (see Appendix~\ref{apx:eval_on_bbl}), the impact of compressor model size (see Appendix~\ref{apx:imp_model_size}), and case study (see Appendix~\ref{apx:case_study}).

\begin{table}[htb]
\centering
\label{tab:comparison}
\setlength{\tabcolsep}{6.5mm}
\resizebox{1.0\columnwidth}{!}{
\begin{tabular}{l|ccc}
\toprule
Methods & F1 & Tokens & 1/$\tau$ \\
\midrule
\multicolumn{4}{@{}l}{{ \textit{Retrieval-based Methods}}} \\ 
SentenceBert & 26.8 & 1,368 & 1.9x \\
BgeReranker & 25.9 & 1,371 & 1.9x \\
BgeLLMembedder & 26.9 & 1,371 & 1.9x \\
\cmidrule (r){1-1}\cmidrule (lr){2-4} 
\multicolumn{4}{@{}l}{{ \textit{Compression-based Methods}}} \\ 
SelectiveContext & 15.8 & 1,553 & 1.7x \\
LLMLingua & 15.2 & 1,343 & 1.9x \\
LLMLingua2 & 21.3 & 1,298 & 2.0x \\
LongLLMLingua & 23.5 & 1,364 & 1.9x \\
\cmidrule (r){1-1}\cmidrule (lr){2-4} 
{\cellcolor[rgb]{0.925,0.957,1}}\textbf{Perception Compressor} & {\cellcolor[rgb]{0.925,0.957,1}}\textbf{29.2} & {\cellcolor[rgb]{0.925,0.957,1}}1,287 & {\cellcolor[rgb]{0.925,0.957,1}}2.0x \\
\midrule
Original Prompt & 25.3 & 2,571 & - \\
\bottomrule
\end{tabular}
    }
\caption{Performance of different methods on MuSicQue~\cite{trivedi2022MuSiQue} with 2x constraint using LLaMA-3-8B-Instruct.}
\label{tab:MuSiQue}
\vspace{-2mm}
\end{table}

\subsection{Datasets \& Evaluation Metric}

We evaluate our method on three long context benchmarks. To study the impact of the position of key information, we utilize NaturalQuestions~\cite{liu2024lost}. For the multi-hop question answering (QA) task, we test on MusicQue~\cite{trivedi2022MuSiQue}. Furthermore, to comprehensively assess the performance of compressed prompts across various long context scenarios, we use the English portion of LongBench~\cite{bai2023longbench}. More details are provided in Appendix~\ref{apx:dataset_details}.


\subsection{Implementation Details}
\label{sec:imp_detail}
In this paper, We use LLaMA-2-7B to compress prompts. We validate the effectiveness of compressed prompts with LongChat-7B-v1.5-32k and LLaMA-3-8B-Instruct. We implement our method based on Pytorch 2.2.2 and HuggingFace Transformers. For hyperparameters, we set the coarse-grained control coefficient to 1.1, the open-book coefficient $\tau_o$ to 0.2, and $\tau_{\text{ins}}$ and $\tau_{\text{q}}$ for instruction and input question to 0.95 and 0.9, respectively. Slopes $k_1$ and $k_2$ are set to 0.4 and 0.1. More details are provided in the Appendix~\ref{apx:imp_details}.

\subsection{Baseline Methods}
We consider the following baseline methods.
\begin{itemize}
    \item \textit{Retrieval-based~Methods}. Retrieval-based methods include several SOTA methods: BM25, Sentence Bert~\cite{reimers2019sentence}, Jina~\cite{günther2023jina}, Gzip~\cite{jiang-etal-2023-low}, OpenAI Embeding\footnote{https://platform.openai.com/docs/guides/embeddings}, Bge, BgeReranker and Bge LLMembedder~\cite{bge_embedding}. In the recall experiment (Table~\ref{tab:recall}), we compare our method with all these methods, but in the compression experiments (Table~\ref{tab:nq}, \ref{tab:longbench} and \ref{tab:MuSiQue}), we select the top four methods with high recall rates for comparison, \textit{i.e.}, SentenceBert, BgeLLMembedder, BgeReranker, and OpenAI Embedding. We discard sentences or tokens with low association with the input question until the compression constraint is met. Due to economic constraints, OpenAI Embedding is only compared in the experiments on NaturalQuestions.
    \item \textit{Compression-based Methods}. Compression-based methods also include four SOTA methods explicitly compressing prompts: SelectiveContext~\cite{li2023compressing}, LLMLingua~\cite{jiang2023llmlingua}, LLMLingua2\footnote{Specifically, we compare with LLMLingua2-large in all experiments.}~\cite{pan2024llmlingua}, and LongLLMLingua~\cite{jiang2023longllmlingua}. As for the LLM for compressing context, SelectiveContext uses the default GPT-2, while LLMLingua and LongLLMLingua use LLaMA-2-7B.
\end{itemize}

\subsection{Results \& Discussions}
We analyze experiment results and illustrate our findings in the context of answering our research questions.

\textbf{1. How does Perception Compressor address the two challenges in long context scenarios, \textit{i.e.}, sensitivity to the position of key information, and excessively long input sequences?}

In long context scenarios, LLMs achieve the highest performance when key information appears at the beginning of the prompt, but if the key information appears in the middle, LLM's performance drops significantly, \textit{i.e.}, \emph{lost in the middle}~\cite{liu2024lost}. To address this challenge, we introduce Perception retriever, which first calculates a perception perplexity $r_k$ for each demonstration, representing the relevance of each demonstration to the input question, and then reorders the demonstrations from $r_k$ in descending order, \textit{i.e.}, the demonstration more relevant to the input question is placed at the front, thus addressing the \emph{lost in the middle} challenge. Perception Compressor has the highest recall@1 in the recall experiment, which proves that it has superior re-ranking performance compared to LongLLMLingua.

As shown in Table~\ref{tab:nq}, Table~\ref{tab:longbench}, and Table~\ref{tab:MuSiQue}, Perception Compressor achieves state-of-the-art (SOTA) performance while drastically reducing the input length. This effectively addresses the challenge of excessively long input sequences with too much redundant information.

\textbf{2. How does Perception Compressor compare to baseline methods in long context benchmarks?}

Table~\ref{tab:recall}, Table~\ref{tab:nq}, Table~\ref{tab:longbench}, and Table~\ref{tab:MuSiQue} present the results of our method compared to baseline methods on NaturalQuestions, LongBench, and MuSiQue. Here are some findings: (1) Perception Compressor outperforms all baseline methods by a large margin across all compression constraint ($1/\tau$) settings and tasks, using nearly the fewest input tokens. This highlights the robustness and superiority of the Perception Compressor. For example, at the 10th position under a 4x compression ratio, Perception Compressor improves by 5.9\% compared to the second-best method, LongLLMLingua, using fewer input tokens. (2) Retrieval-based methods, constrained by relatively low recall, perform well across all datasets but fail to achieve the highest performance. (3) Our Perception retriever surpasses all comparison methods at recall@1-13, with larger gaps as the number of retrieved demonstrations decreases. Notably, at recall@1, it exceeds LongLLMLingua by 5.2\%, showcasing the high upper bound of our method under high compression ratios. (4) Methods that do not consider the input question, such as LLMLingua, SelectiveContext, and LLMLingua2, perform poorly in long context scenarios, clearly inferior to retrieval-based methods. LLMLingua and SelectiveContext even perform worse than zero-shot settings on NaturalQuestions. (5) Even for multi-hop questions, the Perception Compressor remains effective. This indicates that when key information is dispersed, the Perception Compressor can still accurately sense the position of different key information and achieve the best performance.

\textbf{3. How do the individual components of Perception Compressor influence its success?}

To evaluate the contributions of different components in Perception Compressor, we conduct five variants of it for an ablation study (see Table~\ref{tab:ablation}). (1)\textit{Our w/o Perception Retriever} refers to first reordering demonstrations based on their average perplexity from high to low, then retaining demonstrations with higher average perplexity, which may provide more supplemental knowledge to the LLM~\cite{jiang2023llmlingua}.(2)\textit{Our w/o Dual-slope Ratio Allocator} disregards the open-book coefficient $\tau_o$, slopes $k_1$ and $k_2$, directly computing $\tau_{\text{dems}}$ based on pre-defined $\tau$, $\tau_{\text{q}}$, and $\tau_{\text{ins}}$ using Equation~\eqref{eq:cal tau dems}. (3)\textit{Our w/o Semi-guided Iterative Compression} only retains tokens with higher perplexity during the iterative compression process. (4)\textit{Our w LongLLMLingua $r_k$} denotes using LongLLMLingua $r_k$ instead of the Perception retriever $r_k$ during the retrieval stage. (5)\textit{Our w contrast perplexity ITPC} indicates replacing the semi-guided iterative compression with contrast perplexity ITPC, which is the token-level compression algorithm in LongLLMLingua.

As shown in Table~\ref{tab:ablation}, removing any component of the Perception Compressor leads to a significant drop in performance, which well validates the effectiveness of each component. The removal of the perception retriever leads to the most significant performance drop, which may owe to the loss of key information and the \emph{lost in the middle} challenge. We also conduct a comprehensive comparison with LongLLMLingua, and the use of LongLLMLingua $r_k$ and Contrast ITPC both result in inferior performance, which demonstrates that our method is superior to LongLLMLingua.


\begin{table}[ht]
    \centering
     \resizebox{1.0\columnwidth}{!}{
    \begin{tabular}{lcccccc}
    \toprule
         & 1st & 5th & 10th & 15th & 20th & Tokens \\
         \midrule
        \textbf{Perception Compressor} & \textbf{79.5} & \textbf{78.3} & \textbf{78.6} & \textbf{79.1} & \textbf{79.1} & 1,373 \\
        - w/o Perception Retriever & 45.3 & 45.9 & 45.6 & 46.4 & 46.1 & 1,385  \\
        - w/o Semi-guided Iterative Compression & 78.3 & 77.3 & 77.5 & 78.0 & 78.2 & 1,405 \\
        - w/o Dual-slope Ratio Allocator & 76.2 & 76.1 & 76.4 & 76.2 & 76.3 & 1,456 \\
        - w LongLLMLingua $r_k$ & 77.1 & 77.3 & 76.9 & 77.0 & 77.4 & 1,377 \\
        - w contrast perplexity ITPC  & 78.0 & 77.2 & 77.6 & 78.0 & 78.0 & 1,382 \\
        \bottomrule
    \end{tabular}
    }
    \caption{Ablation study on NaturalQuestions under 2x constraint using LLaMA-3-8B-Instruct.}
    \label{tab:ablation}
    \vspace{-1mm} 
\end{table}

\section{Conclusion}
In this paper, we present the Perception Compressor, a training-free prompt compression framework. It consists of three components: a perception retriever, a dual-slope ratio allocator, and a semi-guided iterative compression. Our method compresses prompts via demonstrations reordering, compression ratios and open-book ratios allocation, preservation KITs, and removal of high perplexity NITs. We conduct extensive experiments on LongBench, NaturalQuestions, and MuSiQue. The experiment results demonstrate that Perception Compressor achieves SOTA performance across all tasks, effectively solving two challenges, \textit{i.e.}, \textit{lost in the middle}, and excessively long input sequences. Our method has the potential to enhance LLM performance while substantially compressing long context.

\section*{Limitations}
There are also some limitations in our method. (1) As an explicit prompt compression method, Perception Compressor has a limited upper bound on compression ratio, and obviously cannot achieve the same level of context compression into one token as in \citet{Cheng2024xRAGEC}. (2) If the demonstrations in the context are not well-defined, they need to be divided based on separators or semantics, as there is no fixed algorithm for dividing demonstrations.

\section*{Acknowledgements}
This research is supported by the National Natural Science Foundation of China (Grant No. 62276154), the Research Center for Computer Network (Shenzhen) Ministry of Education, the Natural Science Foundation of Guangdong Province (Grants No. 2023A1515012914 and 440300241033100801770), the Basic Research Fund of Shenzhen City (Grants No. JCYJ20210324120012033, JCYJ20240813112009013, and GJHZ20240218113603006), and the Major Key Project of PCL (Grants No. PCL2022A05 and PCL2023A09).

\bibliography{acl_latex}

\newpage
\appendix

\section{Theoretical Derivation of Perception Perplexity $r_k$}
\label{apx:derivation}
The theoretical derivation of our method primarily relies on Bayes' theorem~\cite{bayes1958essay}.
\subsection{Bayes's Theorem}
 Bayes' Theorem is a mathematical formula used to update the probability of a hypothesis based on new evidence. It is expressed as:

\begin{equation}
   p(b|a) = \frac{p(a|b) \cdot p(b)}{p(a)} 
\end{equation}

Where \( p(b|a) \) is the posterior probability of the hypothesis \(b\) given the evidence \(a\); \( p(a|b) \) is the likelihood of the evidence given that the hypothesis is true; \( p(b) \) is the prior probability of the hypothesis before considering the evidence; \( p(a) \) is the total probability of the evidence across all possible hypotheses.

\subsection{Theoretical Derivation}

\begin{equation}
\label{eq:bayes_qianti1}
\begin{aligned}
    r_{k,j} = &\sum_{i=1}^{L_{ins} + L_{q_j} + L_{r}} g\left( \mathbf{x}^{\text{con}_j}_i \right) \log p\left( \mathbf{x}^{\text{con}_j}_i \mid \mathbf{x}^{\text{dems}_{k}} \right)
\end{aligned}
\end{equation}

where $g(\cdot)$ represents the probability distribution of the ground truth, \textit{i.e.}, $g\left( \mathbf{x}^{\text{con}_j}_i \right)$ is constant; $\mathbf{x}^{\text{con}_j}$ is the concatenation of $\mathbf{x}^{\text{ins}}$, $\mathbf{x}^{\text{q}_{j}}$ and $\mathbf{x}^{\text{r}}$; $\mathbf{x}^{\text{con}_j}_i$ denotes the $i$-th token in $\mathbf{x}^{\text{con}_j}$; $\mathbf{x}^{\text{dems}_{k}}$ and $\mathbf{x}^{\text{r}}$ refer the $k$-th demonstration and regularization constraint\footnote{For fair comparison, We use the same regularization constraint as in \citet{jiang2023longllmlingua}.}, respectively.

According to Bayes' Theorem, we can get:

\begin{equation}
\label{eq:bayes_qianti2}
\small
\begin{aligned}
p\left(\mathbf{x}^{\text{dems}_k} \mid \mathbf{x}^{\text{con}_j}_i \right)&=p\left(\mathbf{x}^{\text{dems}_k}\right) \frac{p\left(\mathbf{x}^{\text{con}_j}_i \mid \mathbf{x}^{\text{dems}_k} \right)}{p\left( \mathbf{x}^{\text{con}_j}_i \right)}
\end{aligned} 
\end{equation}

where $p\left(\mathbf{x}^{\text{dems}_k}\right)$ and $p\left( \mathbf{x}^{\text{con}_j}_i \right)$ are constants.

We can derive from Equation~\eqref{eq:bayes_qianti1} and Equation~\eqref{eq:bayes_qianti2} that:
\begin{equation}
\label{eq:r_k_j_pro}
    \{r_{k,j}\}_i \propto \log p\left(\mathbf{x}^{\text{dems}_k} \mid \mathbf{x}^{\text{con}_j}_i \right)
\end{equation}

where $\{r_{k}\}_i$ is the impact of the $i$-th token in $\mathbf{x}^{\text{con}_j}$ on the perception perplexity. 

\begin{equation}
\label{eq:ws_perception_perlexity_d}
    r_k = \sum_{j=0}^{n} w_j \cdot r_{k,j}
\end{equation}

According to Equation~\eqref{eq:ws_perception_perlexity_d} and Equation~\eqref{eq:r_k_j_pro}, we can derive:

\begin{align*}
\{r_k\}_i &= \sum_{j=0}^{n} w_j \{r_{k,j}\}_i \\
&\propto \sum_{j=0}^{n} w_j \log p\left(\mathbf{x}^{\text{dems}_k} \mid \mathbf{x}^{\text{con}_j}_i \right)
\end{align*}

So, we can get:
\begin{equation}
\small
\begin{aligned}
\{r_k\}_i \propto \log \prod_{j=0}^{n}  {p\left(\mathbf{x}^{\text{dems}_k} \mid \mathbf{x}^{\text{con}_j}_i \right)}^{w_j}
\end{aligned} 
\end{equation}


\section{Parameter Sensitivity Analysis}
\label{apx:psa}
$\tau_{o}, k_1, k_2$ are the three main hyperparameters of the Perception Compressor. To explore the performance of the Perception Compressor under different combinations of $\tau_{o}, k_1, k_2$, we conduct a parameter sensitivity analysis. Specifically, we study the impact of each parameter on performance individually, allowing the value of the parameter under study to range from 0.2 to 0.8 in increments of 0.2 (As mentioned in Appendix~\ref{apx:imp_details}, the initial parameter combination is $k_1=0.4, k_2=0.1, \tau_{o}=0.2$).
\begin{figure}[htbp] 
    \centering 
    \includegraphics[width=1\linewidth]{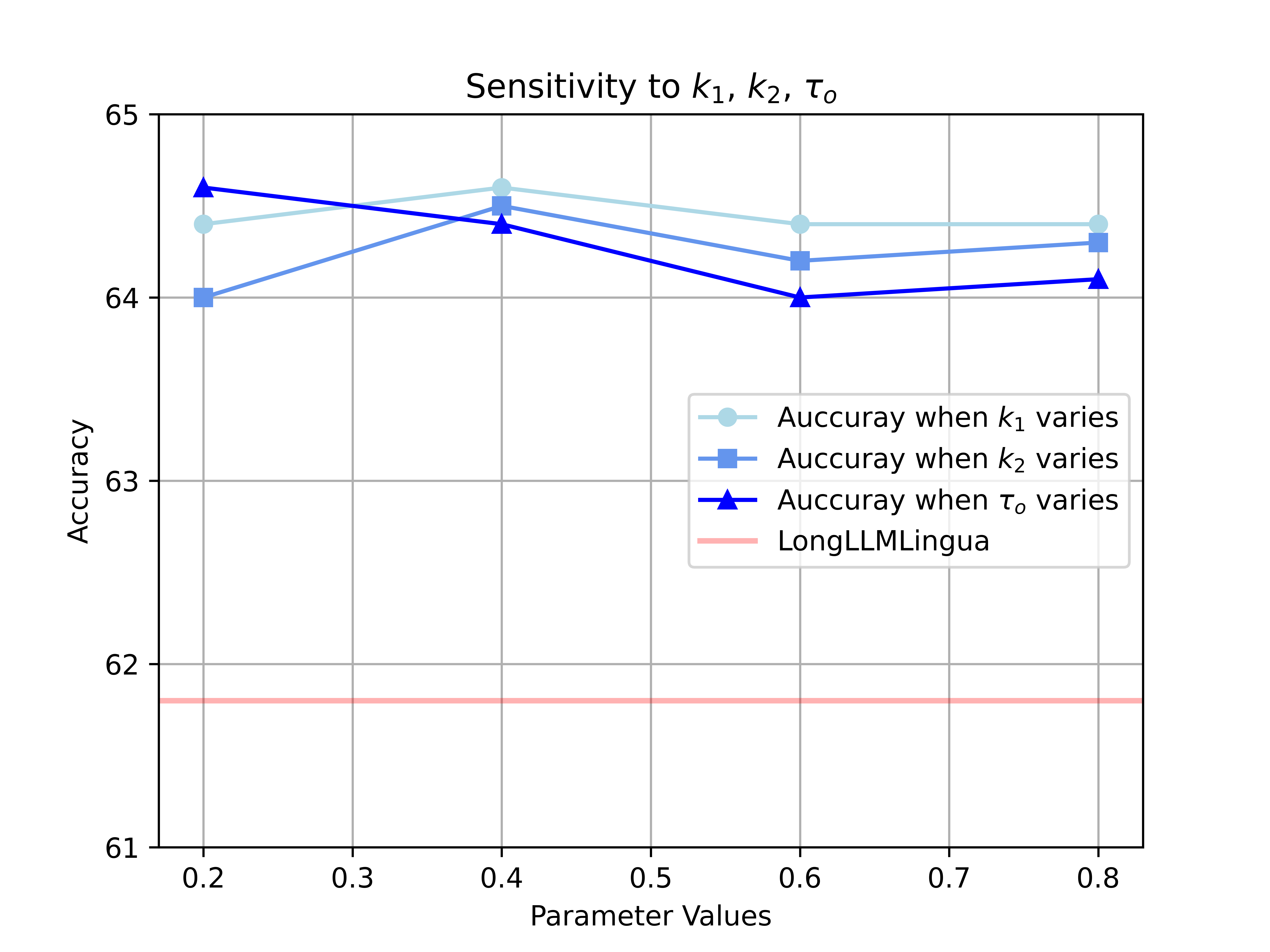} 
    \caption{Parameter Sensitivity Analysis on the 20th position of NatureQuestions under 2x constraint. We use LongChat-7B-v1.5-32k as the response model.} 
    \label{fig:para_ana} 
\end{figure}

The experimental results are shown in Figure~\ref{fig:para_ana}, from which we can draw two findings: (1) The performance of the Perception Compressor is highly stable under different parameter combinations, with fluctuations in Accuracy of less than 1\%, indicating that the Perception Compressor does not require exclusive tuning. (2) Perception Compressor significantly outperforms the previous state-of-the-art method, LongLLMLingua, under all parameter combinations, demonstrating its superiority.

\begin{table*}[thb]
    \centering
     \resizebox{1.0\textwidth}{!}{
    \begin{tabular}{l|cc|cc|cc}
    \toprule
         \multirow{2}{*}{Methods} & \multicolumn{2}{@{}c}{{\bf Memory Usage}} & \multicolumn{2}{@{}c}{{\bf Latency}} & \multicolumn{2}{@{}c}{{\bf Length}} \\
         & Compression Memory Usage & Infer Memory Usage & Compression Latency & Infer Latency & Tokens & $1/\tau$ \\
        \midrule
        \midrule
        \multicolumn{7}{@{}c}{{ \textit{2x constraint}}} \\
        \midrule
        LongLLMLingua & 17,154.9 & 17,627.0 & 1.8 & 2.3 & 1,357 & 2.2x \\
        \cmidrule (r){1-1}\cmidrule (lr){2-5}\cmidrule (lr){6-7} 
        {\cellcolor[rgb]{0.925,0.957,1}}Perception Compressor & {\cellcolor[rgb]{0.925,0.957,1}}18,383.0 & {\cellcolor[rgb]{0.925,0.957,1}}17,634.2 & {\cellcolor[rgb]{0.925,0.957,1}}3.0 & {\cellcolor[rgb]{0.925,0.957,1}}2.3 & {\cellcolor[rgb]{0.925,0.957,1}}1,346 & {\cellcolor[rgb]{0.925,0.957,1}}2.2x \\
        \midrule
        \midrule
        \multicolumn{7}{@{}c}{{ \textit{4x constraint}}} \\
        \midrule
        LongLLMLingua & 15,545.0 & 16,967.3 & 1.0 & 2.2 & 697 & 4.3x \\
        \cmidrule (r){1-1}\cmidrule (lr){2-5}\cmidrule (lr){6-7} 
        {\cellcolor[rgb]{0.925,0.957,1}}Perception Compressor & {\cellcolor[rgb]{0.925,0.957,1}}17,031.0 & {\cellcolor[rgb]{0.925,0.957,1}}16,996.2 & {\cellcolor[rgb]{0.925,0.957,1}}2.2 & {\cellcolor[rgb]{0.925,0.957,1}}2.2 & {\cellcolor[rgb]{0.925,0.957,1}}690 & {\cellcolor[rgb]{0.925,0.957,1}}4.3x \\
        \midrule
        Original Prompt & - & 19,625.8 & - & 2.5 & 2,949 & - \\
        \bottomrule
    \end{tabular}
    }
    \caption{The comparison of Latency (s) and Memory Usage (MB) between Perception Compressor and LongLLMLingua on NatureQuestions. Each value in the table is the average of experimental results from five positions (i.e., 1st, 5th, 10th, 15th, and 20th).}
    \label{tab:comp_lantecy}
\end{table*}

\section{Latency and Memory Usage}
\label{apx:latency_and_mem}
We conduct experiments on Latency and Memory Usage on NatureQuestions using two NVIDIA A800 GPUs. The LLM used for testing Infer Latency is LLaMA-3-8B-Instruct, and other implementation details can be found in Sec.~\ref{sec:imp_detail} and Appendix~\ref{apx:imp_details}.

From Table~\ref{tab:comp_lantecy}, we can draw three findings: (1)Under different compression constraints, the Compression Latency and Compression Memory Usage of Perception Compressor are slightly higher than those of LongLLMLingua. During the compression process, Perception Compressor used additional guiding questions in the retrieval phase, resulting in slightly higher Compression Latency and Compression Memory compared to LongLLMLingua, with Compression Latency increasing by approximately 1.2 seconds and Compression Memory Usage increasing by about 1 GB. Despite the minor cost in time and space, the performance improvement is significant, demonstrating the effectiveness of our method. (2)The Inference Latency of prompts compressed by LongLLMLingua and Perception Compressor is lower than the Latency of directly inputting the Original Prompt, but the reduction in Latency is minimal. This may be because the token length of the Original Prompt is not long to begin with, so the size of the large language model (LLM) itself is the main factor affecting the Inference Latency at this time. (3)The Infer Memory Usage of compressed prompts is significantly reduced. The Infer Memory Usage of compressed prompts is reduced from approximately 19 GB to about 17 GB.

\section{Evaluation on Black-box Large Language Models}
\label{apx:eval_on_bbl}
We conduct further evaluation using the latest black-box large model GPT-4o-mini on NatureQuestions (see Table~\ref{tab:recall}). 

Our findings reveal that Perception Compressor consistently outperforms LongLLMLingua under various compression constraints, with the gap widening as the compression constraint increases. This observation strongly validates the superiority and effectiveness of Perception Compressor. This may be attributed to Perception Compressor's highest recall@1 (see Table~\ref{tab:gpt-4o}), which enhances the likelihood of ground truth documents being ranked at the beginning of prompts and the probability of retaining key information under high compression rates.
\begin{table}[thb]
    \centering
     \resizebox{1.0\columnwidth}{!}{
    \begin{tabular}{l|ccccc|c}
    \toprule
         Methods & 1st & 5th & 10th & 15th & 20th & Tokens \\
        \midrule
        \midrule
        \multicolumn{7}{@{}c}{{ \textit{2x constraint}}} \\
        \midrule
        LongLLMLingua & 76.9 & 77.4 & 77.6 & 77.4 & 77.6 & 1,444 \\
        \cmidrule (r){1-1}\cmidrule (lr){2-7} 
        {\cellcolor[rgb]{0.925,0.957,1}}\textbf{Perception Compressor} & {\cellcolor[rgb]{0.925,0.957,1}}\textbf{78.2} & {\cellcolor[rgb]{0.925,0.957,1}}\textbf{77.6} & {\cellcolor[rgb]{0.925,0.957,1}}\textbf{78.6} & {\cellcolor[rgb]{0.925,0.957,1}}\textbf{78.2} & {\cellcolor[rgb]{0.925,0.957,1}}\textbf{78.5} & {\cellcolor[rgb]{0.925,0.957,1}}\textbf{1,373} \\
        \midrule
        \midrule
        \multicolumn{7}{@{}c}{{ \textit{4x constraint}}} \\
        \midrule
        LongLLMLingua & 77.8 & 76.3 & 76.6 & 76.5 & 76.9 & 736 \\
        \cmidrule (r){1-1}\cmidrule (lr){2-7} 
        {\cellcolor[rgb]{0.925,0.957,1}}\textbf{Perception Compressor} & {\cellcolor[rgb]{0.925,0.957,1}}\textbf{78.9} & {\cellcolor[rgb]{0.925,0.957,1}}\textbf{79.2} & {\cellcolor[rgb]{0.925,0.957,1}}\textbf{79.1} & {\cellcolor[rgb]{0.925,0.957,1}}\textbf{78.5} & {\cellcolor[rgb]{0.925,0.957,1}}\textbf{79.4} & {\cellcolor[rgb]{0.925,0.957,1}}\textbf{697} \\
        \bottomrule
    \end{tabular}
    }
    \caption{Performance Comparison of LongLLMLingua and Perception Compressor on NatureQuestions under 2x constraint. The response model is GPT-4o-mini.}
    \label{tab:gpt-4o}
\end{table}

\section{Implementation Details}
\label{apx:imp_details}
We use LLaMA-3-8B-Instruct tokenizer to count all the tokens. We generate three guiding questions for each input question and set the segment size to 200 tokens. For retrieval-based methods, we use the current strongest baseline model, and the relation between methods and versions is shown in Table~\ref{tab:versions}. We conduct all experiments on 8 NVIDIA GeForce RTX 3090 GPUs.
\begin{table}[ht]
    \small
    \centering
    \begin{tabular}{l|l}
    \toprule
        Methods & Corresponding Version  \\
        \cmidrule (r){1-1}\cmidrule (lr){2-2} 
        OpenAI Embedding & text-embedding-3-large \\
        SentenceBert & all-mpnet-base-v2 \\
        Bge & bge-large-en-v1.5 \\
        BgeReranker & bge-reranker-large \\
        BgeLLMembedder & llm-embedder \\
        Jina & jina-embeddings-v2-base-en \\
        Gzip & - \\
        \bottomrule
    \end{tabular}
    \caption{The relation between methods and versions. There is only one version of Gzip.}
    \label{tab:versions}
\end{table}

\section{Dataset Details}
\label{apx:dataset_details}
\paragraph{NaturalQuestions Multi-document QA} NaturalQuestions Multi-document QA dataset is constructed by~\citet{liu2024lost} based on the NaturalQuestions dataset~\cite{Kwiatkowski_Palomaki_Redfield_Collins_Parikh_Alberti_Epstein_Polosukhin_Devlin_Lee_et_al._2019}, containing 2,655 questions. Each sample in the dataset includes a question and k related documents. These k related documents are obtained from historical queries issued to the Google search engine and human-annotated answers extracted from Wikipedia using the Contriever retrieval system~\cite{izacard2021towards}, with only one of them containing the correct answer document. In our experiments, we used the version with 20 documents, where the dataset includes five different true document position settings in the prompt: 1st, 5th, 10th, 15th, and 20th. Following~\citet{liu2024lost}, we use Accuracy as the evaluation metric. The average prompt tokens length in this benchmark is 2,949.
\paragraph{LongBench}LongBench~\cite{bai2023longbench} is a long context benchmark, and its English portion includes 16 tasks and 3,750 questions. 
\begin{figure}[thb]
    \begin{tcolorbox}
    \textbf{Input Question:} \\
    where did the titanic sink at what ocean? \\
    
    \textbf{Prompt:} \\
    Please provide three most helpful guiding questions to address the original question: where did the titanic sink at what ocean?\\
    
    \textbf{GPT-3.5-Turbo's Response:} \\
    1. Can you recall any historical events related to a famous ship sinking? \\
    2. What do you know about the Titanic and its tragic fate? \\
    3. Can you identify any significant bodies of water where maritime disasters have occurred? \\
    \end{tcolorbox}
    \caption{Generate guiding questions for the input question "where did the titanic sink at what ocean?".}
    \label{fig:generate_guiding_questions}
\end{figure}
These tasks can be divided into six categories, comprehensively covering various long context application scenarios, \textit{i.e.}, multi-document question answering, single-document question answering, few-shot learning, code completion, synthetic tasks, and summarization. In our experiment, we use the English portion of this dataset and the evaluation script provided with it. The average prompt token length in this benchmark is 10,276. 
\paragraph{MuSiQue} MuSiQue~\cite{trivedi2022MuSiQue} dataset is a multi-hop question-answering dataset consisting of 39,876 training samples, 4,834 validation samples, and 4,918 test samples. Among them, the validation set is composed of 2,411 answerable questions and 2,507 unanswerable questions. Completing this task requires large language model (LLM) to conduct multiple inferences based on several documents, necessitating the capability for global information processing. In our experiments, we use the answerable questions from the validation set. Following \citet{trivedi2022MuSiQue}, we report standard F1 as the evaluation metric. The average prompt token length in this benchmark is 2,571.

\section{Cases Study}
\label{apx:case_study}
Large language models (LLMs) effectively comprehend the semantic information in the compressed prompts, even if it might be challenging for humans~\cite{gilbert2023semantic,jiang2023llmlingua}. 

Observing the compressed prompts (see Figure~\ref{fig:case_multi_doc_qa} and Figure~\ref{fig:case_code}), we can find that the key information is placed at the beginning and remains intact. As shown in Figure~\ref{fig:case_multi_doc_qa}, the Document [19] containing the ground truth, which should appear at the end of the prompt, is reordered to the beginning of the prompt, and the key information $\boxed{\text{March 8, 2018}}$ remains intact. As shown in Figure~\ref{fig:case_code}, the function \verb|instructionNames|, which is very similar to the function \verb|fieldNames|, is preserved under a high compression ratio and is reordered to the beginning. The key information \verb|for(InlinedInstruction inst : insts) {| also remains intact. This indicates that Perception Compressor not only removes redundant information and significantly reduces the length of the input sequences but also solves the \textit{lost in the middle} challenge raised in \citet{liu2024lost}, demonstrating its superiority and effectiveness. Therefore, the LLM generates correct responses to the seemingly corrupted compressed prompts.

  
  
  
  


\section{Generate Guiding Questions}
In our experiment, we utilize GPT-3.5-Turbo to generate guiding questions for the input question (see Figure~\ref{fig:generate_guiding_questions}). Generating guiding questions related to the input question can help clarify thoughts and gradually break down the input question. Guiding questions often consider the input question from different angles, uncovering potential factors to help identify underlying assumptions and prerequisites. it is promising to propagate insights of solving guiding questions to inspire solving the input question.

\section{Impact of Model Size}
\label{apx:imp_model_size}
The parameter size of LLMs affects their performance~\cite{kaplan2020scaling}. To explore the impact of parameter scale on performance, we use GPT-2, a small LLM with only 137M parameters, to compress prompt instead of LLaMA-2-7B. As shown in Table~\ref{tab:gpt}, we can draw the following conclusions: (1) The prompt compression effect of GPT-2 is obviously not as good as that of LLaMA-2-7B, which may be due to the larger parameter size of the LLM having stronger capabilities. (2) Compared with other methods, the performance of Perception Compressor using GPT-2 is only lower than LongLLMLingua using LLaMA-2-7B and Original Prompt in the 1st position, while it achieves the highest performance in other settings. This demonstrates the superiority and effectiveness of our method.

\begin{table}[thb]
    \centering
     \resizebox{1.0\columnwidth}{!}{
    \begin{tabular}{l|cccccc}
    \toprule
         Methods & 1st & 5th & 10th & 15th & 20th & Tokens \\
        \cmidrule (r){1-1}\cmidrule (lr){2-7} 
        \textbf{Perception Compressor} & & & & & & \\
        - w LLaMA-2-7B & \textbf{79.5} & \textbf{78.3} & \textbf{78.6} & \textbf{79.1} & \textbf{79.1} & 1,373 \\
        - w GPT2 & 75.7 & 76.0 & 75.4 & 75.7 & 76.2 & 1,342 \\
        \cmidrule (r){1-1}\cmidrule (lr){2-7} 
        OpenAI & 73.0 & 73.5 & 73.0 & 74.1 & 73.6 & 1,408 \\
        SentenceBert & 72.8 & 73.4 & 72.8 & 73.0 & 72.9 & 1,410 \\ 
        BgeReranker & 74.5 & 74.1 & 74.7 & 73.8 & 72.6 & 1,405 \\
        BgeLLMembedder & 74.0 & 73.7 & 74.2 & 74.0 & 74.0 & 1,407 \\
        LongLLMLingua & 77.0 & 76.0 & 74.8 & 75.2 & 74.9 & 1,444 \\
        \cmidrule (r){1-1}\cmidrule (lr){2-7} 
        Original Prompt & 76.6 & 67.5 & 65.8 & 67.4 & 65.6 & 2,949\\
        \bottomrule
    \end{tabular}
    }
    \caption{Impact of model size on NaturalQuestions under 2x constraint using LLaMA-3-8B-Instruct.}
    \label{tab:gpt}
\end{table}

\begin{figure*}[htb]
    \begin{tcolorbox}
    \textbf{Original Prompt:} \\
    \small
Write a high-quality answer for the given question using only the provided search results (some of which might be irrelevant).\\
Document [0](Title:Jessica Jones (season 3))was ordered in April 2018, a month after the second season was released. Filming for the season began by the end of that June, with Ritter making her directorial debut during the season. The season is scheduled to be released in 2019. Star Krysten Ritter directs an episode for the season, marking her directorial debut. On April 12, 2018, a month after the release of the second season, Netflix ordered a third season of "Jessica Jones". With the season order came confirmation that the returning starring cast would include Krysten Ritter as Jessica Jones, Rachael Taylor as Patricia "Trish" Walker,\\
    \textbf{(omitted some tokens here)}\\
    Question: when is season 2 of jessica jones being released\\
    Answer:\\
\begin{flushright}
\textbf{2866 tokens}
\end{flushright}
    \textbf{Compressed Prompt:} \\
  Write high-quality answer for given question using only the provided search results (some of which might be irrelevant).\\19:Jessica Jones (season 2))The season is scheduled to be released on \boxed{\text{March 8, 2018}}.\\Document [1](Title:Jessica Jones (season 1))shortas. second season of "Jessica Jones" was ordered on January 1, 2016. $<$includeonlyinclude$>$ October 201, Marvel and announced that Marvel Television and ABC Studios wouldflix with live action series aroundaredevil, Jessica Jones, Iron Fist, and Luke Cage, leading up to a miniseries based on the Defenders.issa Rosenberg was brought on to showrun theica Jones series, beured as a "-over" from original she had developed in 2010 for ABC. In December 2014, the title was revealed to be "Marvel's A.K.A. Jessica Jones", [1](Title:Jessica Jones (season 1)) 22, 2017, in Region 2 and Region on December , 2016, and in Region 4 on December 7, 2016. Asflix does notalberership numbers for of their original series,phony Group data the on sample size of 1,000 theirones viewinging's. to Symphony, December 2015, of "Jessica Jones"aged 4. million viewers 35-day viewing. The was by Wurtzel, Nal research [Title:essica Jones (season ))Jica Jones ( 2 second season of the American web television seriesJessica Jones on the Marvel Comics of the same name, follows she on case surrounding encounter with Kilgrave in the Marvel Cinematic Universe (MCU), sharing continuity with the films other television series of the franchise. season by Television association with ABC Studios andall Productions, with Melissa Rosenberg as showrunner. Krysten Ritter as,ael Taylor,rie-Anne Moss, Eka Darville [9](Title:Jessica Jones ( series))before? And answer."essica Jones available streaming serviceflix inories is available inra HD4K dynamic range (HDR). The first season inDR its initial release post-productionuxe The season released opposed toizedrage binge-watching for Netflix original series. Disney Consumer Products a line ofater to more adult audience the'sgier tone. Paul Gitter, [](Title:Jessica Jones (season )) in April 2018, a after the season was released. Filming the began end of directorial debut the. season scheduled to be released in 2019 Starrysten Ritter episode,ingial debut 2, 2018, month of season, Netflix third seasonJessica Jonesation include Krysten Ritterica Jones,ael Taylor asriciaTrish" Walker, [3](Title:Jessica Jones ( ))season,andostoo say season201 Netflixs a other "aredevils 2015 However, Netflix confirmed season 2015 announ November 20. The high range its initial release-productionuxe In 201,intic is based on, [](Title:Jessica Jones (season 1))"etherness mostatable trauma the amount of to othering A trailer released on November , 2015ene Leeds ofThe Street Journal" thatilered empowerment to theating' itsere . Region1u-ray Region\\Question: when is season 2 of jessica jones being released\\Answer:
\begin{flushright}
\textbf{702 tokens}
\normalsize
\end{flushright}
    \textbf{LLaMA-3-8B-Instruct's Response:} \\
    The second season of "Jessica Jones" is scheduled to be released on \boxed{\text{March 8, 2018}}. \\

    \textbf{LongChat-7B-v1.5-32k Response:} \\
    The season is scheduled to be released on \boxed{\text{March 8, 2018}}.  \\

    \textbf{Ground Truth:} \\
    March 8, 2018
    \end{tcolorbox}
    \caption{Case study on NaturalQuestions (20 documents)~\cite{liu2024lost} under 4x constraint.}
    \label{fig:case_multi_doc_qa}
\end{figure*}

\begin{figure*}[htb]
    \begin{tcolorbox}
    \textbf{Original Prompt:} \\
    \scriptsize
    \begin{verbatim}
Please complete the code given below.

/******************************************************************************
 * Copyright (c) 2009 - 2015 IBM Corporation.
 * All rights reserved. This program and the accompanying materials
 * are made available under the terms of the Eclipse Public License v1.0
 * which accompanies this distribution, and is available at
 * http://www.eclipse.org/legal/epl-v10.html
 *
 * Contributors:
 *     IBM Corporation - initial API and implementation
    \end{verbatim}
\textbf{(omitted some tokens here)}\\
    \begin{verbatim}
		final Map<IField,String> field2Name = new LinkedHashMap<IField, String>();
		
Next line of code:
    \end{verbatim}
    \normalsize
    \begin{flushright}
    \textbf{4236 tokens}    
    \end{flushright}
    \textbf{Compressed Prompt:} \\
    \scriptsize
    \begin{verbatim}
Please complete the code given below.

	public static Map<Instruction, String> instructionNames(Set<InlinedInstruction> insts) { 
		final Map<Node,String>Names = nodeNames(s.levantMethods(insts));
		final Map<String, List<InlinedInstruction>> name2Inst = new LinkedHashMap<String, List<InlinedInstruction>>();
		final Map<InlinedInstruction, String> inst2Name = new LinkedHashMap<InlinedInstruction, String>();
		
		for(InlinedInstruction inst : insts) { 
			final String mget(inst.cNode());
			 Stringix
			 = inst.Index();
			if (idx._VALUE) { 
				in";
			 else if (idx.MAX_VALUE) { 
				infix = "end";
			} else { 
				 String =.Class().getSimpleNamereplaceAll(" "replaceAll("Instruction", "");
				fix = cname+idx;			
		final name = m + + infix ""; m;			ListlinedInstruction>2Inst.get(name);			 (namednull) { 
				named = new<InlinedInstruction3				2Inst.put(name,
				}	
	for(Entry, List<linedInstruction>> entry2Inst.entrySet()) { 
			finallinedInstruction> =.getValue();			 (namedsize()1) { 
				2Name.putnamed.(.getKey());					linedInstruction :) { 
					final
					assert.empty					Site it.callStack().();					b.append.itr.next()));
	itrNext {
						_" +.r.

		public static Map<IField, String> fieldNames(Set<IField> fields) { 
		final Map<String, List<IField>> name2Field = new LinkedHashMap<String, List<IField>>();
		final Map<IField,String> field2Name = new LinkedHashMap<IField, String>();
		
Next line of code:
    \end{verbatim}
    \normalsize
\begin{flushright}
\textbf{389 tokens}
\end{flushright}
    \textbf{LLaMA-3-8B-Instruct's Response:} \\
    \scriptsize
    \verb|for(IField field : fields) {| \\
    \normalsize
    \textbf{Ground Truth:} \\
    \scriptsize
    \verb|for(IField field : fields) {| \\
    \normalsize
    \end{tcolorbox}
    \caption{Case study on lcc task in LongBench~\cite{bai2023longbench} under 500 tokens constraint.}
    \label{fig:case_code}
\end{figure*}

\end{document}